\documentclass[letterpaper,10pt,conference]{ieeeconf}

\usepackage{amsmath,amsfonts,amssymb,mathrsfs,bm}
\usepackage{algorithm}
\usepackage{algorithmicx}
\usepackage{algpseudocode}
\usepackage{array}
\usepackage{booktabs}
\usepackage{bbding}
\usepackage{cite}
\usepackage{color,xcolor}
\usepackage{colortbl}
\usepackage{epstopdf}
\usepackage{epsfig}
\usepackage{float}
\usepackage{graphicx}
\usepackage[caption=false,font=normalsize,labelfont=sf,textfont=sf]{subfig}
\usepackage{textcomp}
\usepackage{graphics}
\usepackage{listings}
\usepackage{longtable}
\usepackage{mathptmx}
\usepackage{multirow}
\usepackage{rotating}
\usepackage{stfloats}
\usepackage[mathscr]{euscript}
\usepackage{tikz}
\usepackage{times}
\usepackage{ulem}
\usepackage{url}
\usepackage{verbatim}

\newtheorem{hypothesis}{Hypothesis}

\definecolor{mygray}{gray}{.85}

\newcommand{\M}[1]{\mathbf{#1}}

\IEEEoverridecommandlockouts
\title{\LARGE \bf Diff-PCR: Diffusion-Based Correspondence Search in Doubly Stochastic Matrix Space for Point Cloud Registration}

\author{Haihua Shi, Qianliang Wu
    \thanks{Haihua Shi is with Jinling Institute of Technology, Nanjing, People’s Republic of China. Email: shihaihua@jit.edu.cn}
    \thanks{Qianliang Wu is with PCA Lab, Key Lab of Intelligent Perception and Systems for High-Dimensional Information of Ministry of Education, and Jiangsu Key Lab of Image and Video Understanding for Social Security, School of Computer Science and Engineering, Nanjing University of Science and Technology, Nanjing, People’s Republic of China. Email: wuqianliang@njust.edu.cn}
}

\begin{document}

\maketitle
\thispagestyle{empty}
\pagestyle{empty}

\begin{abstract}
Efficiently identifying accurate correspondences between point clouds is crucial for both rigid and non-rigid point cloud registration. Existing methods usually rely on geometric or semantic feature embeddings to establish correspondences and then estimate transformations or flow fields. Recently, several state-of-the-art methods have adopted RAFT-like iterative updates to refine solutions. However, these methods still have two major limitations. First, their iterative refinement mechanism lacks transparency, and the update trajectory is largely fixed once the refinement starts, which may lead to suboptimal solutions. Second, they overlook the importance of explicitly refining the correspondence matrix before solving for transformations or flow fields. Most existing approaches compute candidate correspondences in feature space and project the resulting matching matrix only once by using Sinkhorn or dual-softmax normalization. Such a one-shot projection can be far from the globally optimal solution, and these methods usually do not model the distribution of the target matching matrix.

In this paper, we propose a novel framework that exploits a denoising diffusion model to predict a search gradient for the optimal matching matrix in doubly stochastic matrix space. Specifically, the diffusion model learns a denoising direction, and the reverse denoising process iteratively searches for improved solutions along this learned direction, which approximates the maximum-likelihood direction of the target matching matrix. To improve efficiency, we design a lightweight denoising module and adopt the accelerated sampling strategy of the Denoising Diffusion Implicit Model (DDIM)\cite{song2020denoising}. Experimental results on 3DMatch/3DLoMatch and 4DMatch/4DLoMatch demonstrate the effectiveness of the proposed framework.
\end{abstract}

\section{Introduction}
Matching point clouds captured from different scans is a fundamental task in many computer vision applications, including point cloud registration\cite{huang2021comprehensive}, scene flow estimation\cite{shen2023self}, and localization\cite{zhang2014loam}. These applications often involve rigid transformations or non-rigid deformations and therefore require accurate point correspondences.

Learning-based methods\cite{qin2022geometric,li2022lepard,yew2022regtr,wu2023sgfeat,mei2023unsupervised} have made remarkable progress in point cloud registration. Many of them employ backbone networks such as KPConv\cite{thomas2019kpconv} to obtain subsampled superpoints and their associated features. These methods typically construct an initial matching matrix in feature space. Outlier rejection modules\cite{bai2021pointdsc,chen2022sc2} can further improve the quality of candidate correspondences by leveraging specially designed networks\cite{qi2019deep,qin2023deep,jiang2023robust} or geometric priors\cite{zhang20233dmac}. Nevertheless, most of these methods still rely on one-shot correspondence prediction, leaving room for further improvement.

Recent works\cite{yu2023peal,li2022lepard,shen2023self,gu2022rcp} have incorporated RAFT-like\cite{teed2020raft} iterative refinement and reported substantial performance gains. However, these methods do not provide an explicit theoretical explanation for why only a small number of refinement steps is usually beneficial. In practice, increasing the number of iterations often degrades performance. Some approaches\cite{mei2023unsupervised,mei2021cotreg,wugraph} attempt to introduce more explicit optimization procedures, but the search process may still become trapped in suboptimal solutions or be driven by infeasible gradient directions because the solution space is highly complex. Moreover, these methods do not explicitly learn the search gradient from data priors encoded by the feature backbone.

Inspired by the reverse sampling process of DDPM\cite{ho2020denoising}, the conditional gradient in the Frank-Wolfe algorithm\cite{jaggi2013revisiting}, and score-based MCMC methods such as Langevin MCMC\cite{parisi1981correlation} and Hamiltonian Monte Carlo\cite{neal2011mcmc}, we propose a robust framework for optimizing the matching matrix in doubly stochastic matrix space\cite{86bd1ad6-50bb-38d0-9978-0966b4dfc6d3}. We argue that optimization should be performed within the feasible solution space and that the search gradient should be learned explicitly by the network rather than being implicitly encoded by RAFT-like iterations\cite{teed2020raft,teed2021raft}. By learning this gradient, the iterative optimization process can move toward the globally optimal matching matrix while being less sensitive to initialization.

Our framework adapts Gaussian reverse sampling from DDPM\cite{ho2020denoising} to iteratively optimize the matching matrix starting from an initial solution. We treat the doubly stochastic matrix space as a continuous relaxation of the discrete correspondence space and therefore apply Gaussian denoising in this continuous domain. During reverse sampling, Sinkhorn normalization is used to enforce the matrix constraints. Unlike deterministic iterative optimization, our approach introduces stochasticity during sampling, which helps the optimizer avoid poor local minima and explore the solution space more effectively. As a result, the learned search process is more robust and supports more refinement steps.

To further improve efficiency, we employ the DDIM\cite{song2020denoising} schedule to accelerate sampling. Our model consists of a KPConv\cite{thomas2019kpconv} backbone and a lightweight denoising module. Extensive experiments show that this lightweight design preserves strong performance while maintaining high efficiency.

Our contributions are summarized as follows:
\begin{itemize}
    \item To the best of our knowledge, this is the first work that applies a diffusion model in doubly stochastic matrix space to iteratively search for an optimal matching matrix through reverse sampling. The proposed matching-matrix diffusion model can potentially be extended to 2D-2D, 2D-3D, and 3D-3D registration problems.
    \item We design a lightweight reverse denoising module that yields fast convergence during iterative optimization. The framework supports both noise-to-target reverse denoising and refinement from a reliable initial matching matrix produced by the backbone.
    \item We conduct extensive experiments on 3DMatch/3DLoMatch and 4DMatch/4DLoMatch. The results demonstrate the effectiveness of the proposed method for both rigid and non-rigid registration and show competitive performance against state-of-the-art approaches.
\end{itemize}

\section{Related Work}

\subsection{Point Cloud Registration}
Recently, feature learning-based point cloud registration methods have achieved significant improvements. Many of them\cite{bai2020d3feat,huang2021predator,yu2021cofinet,qin2022geometric,yu2023rotation,yu2022riga,yu2023peal} rely on the KPConv\cite{thomas2019kpconv} backbone to generate superpoints and associated features with large receptive fields. To further improve performance, these methods often incorporate prior knowledge and design learnable outlier rejection modules. For example, RoITr\cite{yu2023rotation} injects local PPF\cite{deng2018ppf} features into a transformer to enhance rotation invariance. PEAL\cite{yu2023peal} uses a pre-trained registration model to provide overlap information as an overlap prior and then applies a simple attention mechanism across overlapping and non-overlapping regions. Subsequently, a GeoTR\cite{qin2022geometric} network is used for iterative refinement.

Another line of research focuses on outlier rejection among candidate correspondences. PointDSC\cite{bai2021pointdsc} exploits a max-clique algorithm in local patches to cluster inlier correspondences. SC2-PCR\cite{chen2022sc2} constructs a second-order consistency graph for candidate correspondences and provides theoretical guarantees for its robustness. Based on this graph, MAC\cite{zhang20233dmac} develops a variant of the max-clique algorithm to identify more reliable inlier correspondences. In addition, methods such as PEAL\cite{yu2023peal} and DiffusionPCR\cite{chen2023diffusionpcr} employ iterative refinement to improve overlap priors obtained from pre-trained registration models.

\subsection{Diffusion Models for Registration}
Diffusion models\cite{ho2020denoising,song2019generative,song2020denoising} have recently achieved remarkable success in many areas, including object detection\cite{chen2023diffusiondet}, image generation\cite{austin2021structured}, and text-to-image translation\cite{gu2022vector}. These advances stem from either a generative Markov chain based on Langevin MCMC\cite{parisi1981correlation} or a reversed diffusion process\cite{song2020denoising}. Because diffusion models can progressively recover target data distributions from white noise, researchers have begun to apply them to point cloud registration and 6D pose estimation.

A pioneering work\cite{urain2023se} applied a diffusion model in SE(3) space by using NCSN\cite{song2019generative} to learn a denoising score function, which was then used together with Langevin MCMC for 6-DoF grasp pose generation. In addition, \cite{jiang2023se} implemented DDPM\cite{ho2020denoising} in SE(3) space for 6D pose estimation based on a surrogate point cloud registration baseline. Similarly, DiffusionPCR\cite{chen2023diffusionpcr} employs GeoTR\cite{qin2022geometric} as a denoising module to progressively refine overlap priors produced by a pre-trained model, following a strategy related to PEAL\cite{yu2023peal}. In contrast, our method performs diffusion directly in doubly stochastic matrix space, which is applicable to both rigid and deformable registration. By explicitly modeling the constraints of the matching matrix in this space, our method is better suited to handling ambiguities caused by symmetry or repeated structures.

\section{The Proposed Approach}

\subsection{Problem Formulation}
Given a source point cloud $P \in \mathbb{R}^{N \times 3}$ and a target point cloud $Q \in \mathbb{R}^{M \times 3}$, the registration task aims to identify top-$K$ correspondences from a matching matrix $E$ and then estimate a warping function to align $P$ and $Q$. In the rigid case, the warping function $W$ is parameterized by a rigid transformation $T \in \mathrm{SE}(3)$. In the deformable case, the warping can be modeled as point-wise flow fields from $P$ to $Q$. Given the ground-truth warping operation $W_{gt}$, for every correspondence $(p_i \in P, q_j \in Q)$, the constraint
\begin{equation}
    \|W_{gt}(p_i)-q_j\|_2 < \sigma
\end{equation}
should be satisfied, where $\sigma$ is a distance threshold and $\|\cdot\|_2$ denotes the Euclidean norm.

\subsection{Preliminaries}

\textbf{Diffusion models.}
Denoising diffusion probabilistic models (DDPMs)\cite{ho2020denoising,song2020denoising,song2019generative} define a forward noising process that gradually perturbs a clean sample $x_0$ into Gaussian noise, together with a reverse denoising process that recovers $x_0$ from noisy observations. Let $\{\beta_t\}_{t=1}^{T}$ be a variance schedule with $\beta_t \in (0,1)$, and define
\begin{equation}
    \alpha_t = 1-\beta_t, \qquad \bar{\alpha}_t = \prod_{s=1}^{t}\alpha_s .
\end{equation}
The forward Markov process is defined as
\begin{equation}
    q(x_{1:T}\mid x_0)=\prod_{t=1}^{T} q(x_t\mid x_{t-1}),
\end{equation}
with Gaussian transition kernel
\begin{equation}
    q(x_t\mid x_{t-1})=\mathcal{N}\!\left(x_t;\sqrt{\alpha_t}\,x_{t-1},(1-\alpha_t)\mathbf{I}\right).
    \label{diff_3_new}
\end{equation}
By recursively applying Eqn.~(\ref{diff_3_new}), we obtain the closed-form marginal
\begin{equation}
    q(x_t\mid x_0)=\mathcal{N}\!\left(x_t;\sqrt{\bar{\alpha}_t}\,x_0,(1-\bar{\alpha}_t)\mathbf{I}\right).
    \label{diff_4_new}
\end{equation}
Equivalently, a sample at timestep $t$ can be written as
\begin{equation}
    x_t=\sqrt{\bar{\alpha}_t}\,x_0+\sqrt{1-\bar{\alpha}_t}\,\epsilon,
    \qquad
    \epsilon\sim\mathcal{N}(0,\mathbf{I}).
    \label{close_diff_new}
\end{equation}

The posterior distribution of one reverse step conditioned on the clean sample $x_0$ is also Gaussian:
\begin{equation}
    q(x_{t-1}\mid x_t,x_0)=\mathcal{N}\!\left(x_{t-1};\mu_q(x_t,x_0),\Sigma_q(t)\right),
    \label{reverse_trans_new}
\end{equation}
where
\begin{equation}
    \mu_q(x_t,x_0)=
    \frac{\sqrt{\alpha_t}(1-\bar{\alpha}_{t-1})}{1-\bar{\alpha}_t}x_t
    +
    \frac{\sqrt{\bar{\alpha}_{t-1}}(1-\alpha_t)}{1-\bar{\alpha}_t}x_0,
\end{equation}
and
\begin{equation}
    \Sigma_q(t)=\frac{(1-\alpha_t)(1-\bar{\alpha}_{t-1})}{1-\bar{\alpha}_t}\mathbf{I}.
\end{equation}

DDPM learns a parameterized reverse process
\begin{equation}
    p_\theta(x_{0:T})=p(x_T)\prod_{t=1}^{T}p_\theta(x_{t-1}\mid x_t),
\end{equation}
where $p(x_T)=\mathcal{N}(0,\mathbf{I})$. Training is performed by maximizing a variational lower bound. The corresponding ELBO can be written as
\begin{eqnarray}
\begin{aligned}
    \log p_\theta(x_0)
    \geq &
    \ \mathbb{E}_{q}\!\left[\log p_\theta(x_0\mid x_1)\right]
    -D_{KL}\!\left(q(x_T\mid x_0)\,\|\,p(x_T)\right)
    \\
    &
    -\sum_{t=2}^{T}\mathbb{E}_{q(x_t\mid x_0)}
    \left[
    D_{KL}\!\left(
    q(x_{t-1}\mid x_t,x_0)\,\|\,p_\theta(x_{t-1}\mid x_t)
    \right)
    \right].
\end{aligned}
\label{LVB_new}
\end{eqnarray}

A common parameterization is to let the denoising network predict either the injected noise $\epsilon$ or the clean sample $x_0$. In this paper, we adopt the $x_0$-prediction parameterization and use a denoising model $g_\theta(x_t,t)$ to estimate the clean target:
\begin{equation}
    \hat{x}_0=g_\theta(x_t,t).
\end{equation}
Under the Gaussian assumption in Eqn.~(\ref{reverse_trans_new}), minimizing the KL term in Eqn.~(\ref{LVB_new}) is equivalent, up to timestep-dependent constants, to minimizing a weighted mean squared error between the prediction and the clean target:
\begin{equation}
    \mathcal{L}_{x_0}
    =
    \mathbb{E}_{t,x_0,\epsilon}
    \left[
    w_t \left\| g_\theta(x_t,t)-x_0 \right\|_2^2
    \right],
    \label{x0_loss_new}
\end{equation}
where $x_t$ is generated by Eqn.~(\ref{close_diff_new}) and $w_t$ is a non-negative timestep-dependent weight.

\textbf{Doubly stochastic matrix space.}
Let $P\in\mathbb{R}^{N\times 3}$ and $Q\in\mathbb{R}^{M\times 3}$ denote the source and target point clouds. Their correspondence can be represented by a non-negative matrix $E\in\mathbb{R}_{+}^{N\times M}$, where $E_{ij}$ indicates the matching confidence between $p_i\in P$ and $q_j\in Q$. In the ideal one-to-one full matching setting, $E$ reduces to a permutation matrix when $N=M$. In practice, point cloud registration usually involves partial overlap and varying point numbers. Therefore, we relax the discrete correspondence matrix to a non-negative rectangular matrix with approximately normalized row and column marginals.

Following the common practice in matching and optimal transport, we use Sinkhorn normalization\cite{cuturi2013sinkhorn} to project a non-negative score matrix into a relaxed doubly stochastic space:
\begin{equation}
    \mathcal{M}
    =
    \left\{
    A\in\mathbb{R}_{+}^{N\times M}
    \ \middle|\
    A\mathbf{1}_M \approx r,\ 
    A^\top \mathbf{1}_N \approx c
    \right\},
\end{equation}
where $r\in\mathbb{R}_{+}^{N}$ and $c\in\mathbb{R}_{+}^{M}$ are prescribed marginal vectors. In our implementation, $\mathcal{M}$ is realized by a relaxed Sinkhorn projection, which is more suitable than a strict permutation constraint for partial-overlap registration. This continuous relaxation allows us to optimize matching scores while preserving approximate one-to-one structure.

The advantage of searching in the matching-matrix space is that it can directly represent ambiguity caused by symmetry, repeated structures, noise, and non-rigid deformation. Compared with low-dimensional transformation spaces such as $\mathrm{SE}(3)$ or per-point flow spaces, the matching-matrix space provides a richer optimization domain for correspondence reasoning. This motivates our diffusion-based search strategy in $\mathcal{M}$.

\subsection{Diffusion Model in Doubly Stochastic Matrix Space}\label{continuous_diffusion}

We now instantiate the diffusion model for point cloud correspondence estimation. Let $E^0\in\mathcal{M}$ denote the target matching matrix, obtained from ground-truth correspondences or their relaxed supervision. Our goal is to learn a reverse denoising process that gradually recovers $E^0$ from a noisy matrix.

\textbf{Forward diffusion process.}
We define a forward Gaussian noising process over matrices:
\begin{equation}
    q(E^{1:T}\mid E^0)=\prod_{t=1}^{T} q(E^t\mid E^{t-1}),
\end{equation}
with transition kernel
\begin{equation}
    q(E^t\mid E^{t-1})
    =
    \mathcal{N}\!\left(
    E^t;
    \sqrt{\alpha_t}\,E^{t-1},
    (1-\alpha_t)\mathbf{I}
    \right),
    \label{matrix_forward_step}
\end{equation}
where $\mathbf{I}$ denotes the identity covariance over vectorized matrix entries. Therefore, the marginal distribution at timestep $t$ has the closed form
\begin{equation}
    q(E^t\mid E^0)
    =
    \mathcal{N}\!\left(
    E^t;
    \sqrt{\bar{\alpha}_t}\,E^0,
    (1-\bar{\alpha}_t)\mathbf{I}
    \right).
    \label{matrix_forward_closed}
\end{equation}
Equivalently,
\begin{equation}
    E^t=\sqrt{\bar{\alpha}_t}\,E^0+\sqrt{1-\bar{\alpha}_t}\,\epsilon,
    \qquad
    \epsilon\sim\mathcal{N}(0,\mathbf{I}).
    \label{matrix_reparam}
\end{equation}

The matrix sampled from Eqn.~(\ref{matrix_forward_closed}) lives in $\mathbb{R}^{N\times M}$ and does not necessarily satisfy the relaxed doubly stochastic constraints. Instead of enforcing the constraint inside the Gaussian transition itself, we treat $\mathcal{M}$ as the target manifold and allow the intermediate noisy variables to evolve in the ambient Euclidean space. During denoising, the network maps these unconstrained intermediate matrices back toward feasible correspondence matrices in $\mathcal{M}$. This design avoids the difficulty of defining Gaussian diffusion directly on a constrained manifold and preserves the simplicity of standard DDPM parameterization.

\textbf{Reverse denoising process.}
We learn a denoising model $g_\theta(E^t,t,\mathcal{C})$ that predicts the clean matching matrix $E^0$ from a noisy matrix $E^t$, where $\mathcal{C}$ denotes the conditioning information produced by the point cloud backbone, including point coordinates and point features. Formally,
\begin{equation}
    \hat{E}_0 = g_\theta(E^t,t,\mathcal{C}).
    \label{pred_clean_matrix}
\end{equation}
In our framework, the output $\hat{E}_0$ is further regularized by Sinkhorn projection so that it lies in the relaxed doubly stochastic space.

Given the clean prediction $\hat{E}_0$, we adopt the DDIM-style reverse update\cite{song2020denoising}. First, we estimate the noise component as
\begin{equation}
    \hat{\epsilon}_t
    =
    \frac{E^t-\sqrt{\bar{\alpha}_t}\,\hat{E}_0}{\sqrt{1-\bar{\alpha}_t}}.
    \label{eps_est_new}
\end{equation}
Then the previous latent variable is updated by
\begin{equation}
    E^{t-1}
    =
    \sqrt{\bar{\alpha}_{t-1}}\,\hat{E}_0
    +
    \sqrt{1-\bar{\alpha}_{t-1}-\sigma_t^2}\,\hat{\epsilon}_t
    +
    \sigma_t z_t,
    \label{ddim_sampling_new}
\end{equation}
where $z_t\sim\mathcal{N}(0,\mathbf{I})$ for $t>1$ and $z_1=\mathbf{0}$. The noise scale is defined as
\begin{equation}
    \sigma_t
    =
    \eta
    \sqrt{
    \frac{1-\bar{\alpha}_{t-1}}{1-\bar{\alpha}_t}
    }
    \sqrt{
    1-\frac{\bar{\alpha}_t}{\bar{\alpha}_{t-1}}
    },
    \label{sigma_t_new}
\end{equation}
where $\eta\in[0,1]$ controls the trade-off between deterministic sampling and stochastic sampling. When $\eta=0$, Eqn.~(\ref{ddim_sampling_new}) reduces to deterministic DDIM sampling.

Eqns.~(\ref{eps_est_new})-(\ref{sigma_t_new}) are defined in the ambient Euclidean space $\mathbb{R}^{N\times M}$. After each reverse step, the predicted clean matrix $\hat{E}_0$ is projected by Sinkhorn normalization before it is used as the matching matrix for downstream correspondence extraction. This strategy preserves the numerical simplicity of DDIM while maintaining feasibility of the final correspondence matrix.

\textbf{Training objective.}
In principle, one can train $g_\theta$ by the weighted MSE objective in Eqn.~(\ref{x0_loss_new}). However, in correspondence estimation, the target matrix is highly sparse because only a small fraction of entries correspond to true matches. To better address this label imbalance, we supervise the predicted clean matrix by a classification-style loss on matrix entries. Specifically, let $\hat{E}_0=g_\theta(E^t,t,\mathcal{C})$ denote the predicted clean matching matrix after Sinkhorn projection, and let $E^0$ denote the target matching matrix. We optimize
\begin{equation}
    \mathcal{L}_{simple}
    =
    \mathbb{E}_{t,E^0,\epsilon}
    \left[
    \ell\!\left(\hat{E}_0,E^0\right)
    \right],
    \label{L_simple_new}
\end{equation}
where $\ell(\cdot,\cdot)$ is implemented as focal loss in our experiments. This objective can be interpreted as a practical surrogate for the $x_0$-prediction objective in Eqn.~(\ref{x0_loss_new}), adapted to the extreme sparsity of matching supervision.

\textbf{Sampling strategy.}
During inference, we consider two initialization strategies for reverse sampling. The first initializes $E^T$ from Gaussian noise, namely $E^T\sim\mathcal{N}(0,\mathbf{I})$. The second initializes $E^T$ from a reliable matching matrix predicted by the backbone. In both cases, reverse denoising is performed by Eqn.~(\ref{ddim_sampling_new}). To accelerate inference, we use a subsequence of timesteps $\tau=\{\tau_1,\ldots,\tau_S\}\subseteq\{1,\ldots,T\}$ rather than all $T$ steps. This yields a fast DDIM-style sampler that is compatible with both noise initialization and backbone initialization.

\textbf{Remarks.}
Our formulation performs Gaussian diffusion in the ambient Euclidean space and uses Sinkhorn normalization to recover a feasible relaxed matching matrix. This design is intentionally pragmatic. It preserves the simplicity of standard diffusion formulations while incorporating the structural prior of doubly stochastic matching. Empirically, we find that this combination is more stable and easier to optimize than directly defining a constrained stochastic process on the matrix manifold.

\begin{hypothesis}
If there exists a network that can accurately predict the matching score $E_{ij}$ between $p_i$ and $q_j$, and if there exists an efficient optimization algorithm that can search in matrix space, then searching for the optimal solution in the matching-matrix space may yield better results than directly searching in low-dimensional spaces such as $\mathrm{SE}(3)$ and $\mathbb{R}^{3N}/\mathbb{R}^{2N}$.
\end{hypothesis}

\begin{figure*}[htbp]
      \centering
      \includegraphics[width=14cm,height=10cm]{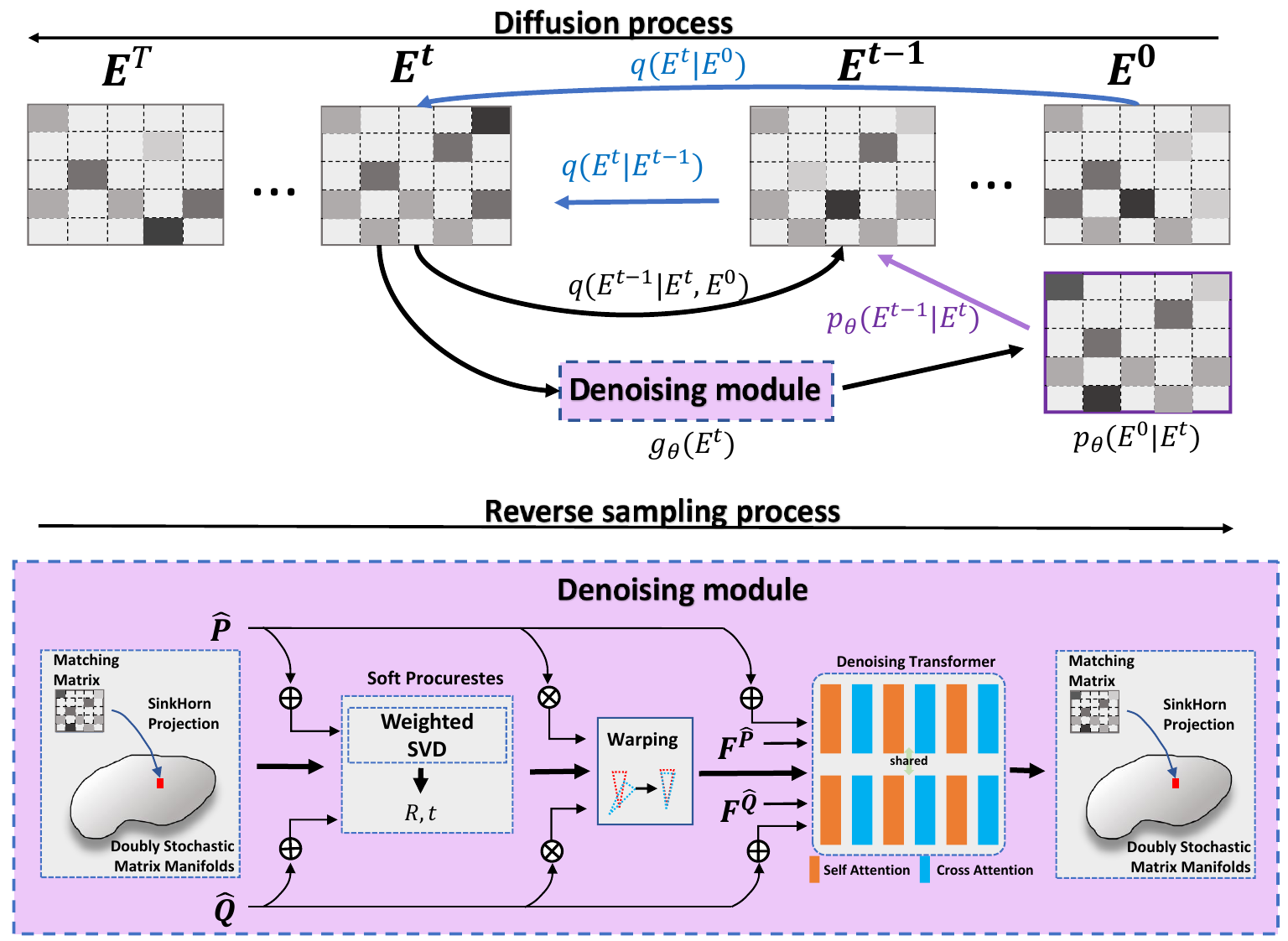}
      \caption{Overview of the proposed matching-matrix diffusion framework. $\bigoplus$ indicates that both 3D coordinates and positional encodings are used as input. $\bigotimes$ indicates that only 3D coordinates are used. After the KPConv backbone extracts the downsampled superpoints $\hat{P}, \hat{Q}$ and their features $F^{\hat{P}}, F^{\hat{Q}}$, these quantities remain fixed during reverse sampling. At each denoising step, they are processed by the warping operation and the denoising transformer. The forward diffusion process is modeled by the Gaussian transition kernel $q(E^t|E^{t-1})$, which has the closed form $q(E^t|E^0)$. The denoising model $g_\theta(E^t)$ learns a reverse denoising direction toward the target solution $E^0$. During inference, we use the predicted $\hat{E}_0$ together with Eqns.~(\ref{matrix_reparam}) and (\ref{ddim_sampling_new}) to sample $E^{t-1}$.}
      \label{framework}
\end{figure*}

\subsection{Framework Overview}
Our framework consists of a KPConv\cite{thomas2019kpconv} feature backbone and a diffusion model\cite{ho2020denoising}. The backbone takes the source point cloud $P$ and target point cloud $Q$ as input and performs three downsampling stages and two upsampling stages to generate downsampled superpoints $\hat{P}$ and $\hat{Q}$, together with their associated features $F^{\hat{P}} \in \mathbb{R}^{N \times d}$ and $F^{\hat{Q}} \in \mathbb{R}^{M \times d}$. The initial matching matrix $E^T$ is either sampled from Gaussian noise or computed from the inner product between $F^{\hat{P}}$ and $F^{\hat{Q}}$. We then apply the denoising module $g_\theta$ to reverse-sample the target matching matrix $E^0$.

The denoising module $g_\theta$ includes four key components:
\begin{itemize}
    \item Sinkhorn projection $\M{f_{sinkhorn}(\cdot)}$, which projects the input matrix into doubly stochastic matrix space.
    \item Weighted SVD $\M{soft\_procrustes(\cdot,\cdot,\cdot)}$, which estimates the rigid transformation $R$ and $t$.
    \item A denoising transformer $f_\theta(\cdot,\cdot,\cdot,\cdot,\cdot,\cdot)$, which updates denoised point features during reverse sampling.
    \item A matching function $\M{matching\_logits(\cdot,\cdot,\cdot,\cdot)}$, which computes the matching matrix between $\hat{P}$ and $\hat{Q}$.
\end{itemize}

\subsection{The Lightweight Denoising Module $g_\theta$}\label{dmd}
In this section, we describe the lightweight denoising module $g_\theta$. We first extract the downsampled superpoints $\hat{P}$ and $\hat{Q}$ and their corresponding features $F^{\hat{P}}$ and $F^{\hat{Q}}$. These quantities remain fixed as input to $g_\theta$ during reverse sampling. The module takes a noisy matching matrix $E^t$ as input and predicts the clean matching matrix $\hat{E}_0$.

Specifically, the denoising module consists of five differentiable components applied sequentially: Sinkhorn projection, weighted SVD, warping, denoising transformer, and matching.

\subsubsection{\uline{Sinkhorn Projection}: $\M{f_{sinkhorn}(\cdot)}$}\label{sinkhorn_projection}
To constrain the matching matrix $E^t$ to the relaxed doubly stochastic space, we use Sinkhorn iterations\cite{cuturi2013sinkhorn} to project $E^t$. In our framework, this is a core component rather than a simple post-processing step.

\subsubsection{\uline{Weighted SVD}: $\M{soft\_procrustes(\cdot,\cdot,\cdot)}$}\label{procrustes}
Given top-$K$ confident correspondences, we use a differentiable weighted SVD algorithm\cite{arun1987least} to estimate the rigid transformation $(R,t)$. Let $\omega_{ij}=\tilde{E}(i,j)$ be the weight of correspondence $(i,j)\in K$. We first compute weighted centroids
\begin{equation}
    \bar{p}=\frac{1}{\sum_{(i,j)\in K}\omega_{ij}}\sum_{(i,j)\in K}\omega_{ij}\hat{p}_i,
    \qquad
    \bar{q}=\frac{1}{\sum_{(i,j)\in K}\omega_{ij}}\sum_{(i,j)\in K}\omega_{ij}\hat{q}_j.
\end{equation}
Then the weighted cross-covariance matrix is
\begin{equation}
    H=\sum_{(i,j)\in K}\omega_{ij}(\hat{p}_i-\bar{p})(\hat{q}_j-\bar{q})^T.
\end{equation}
Let the singular value decomposition of $H$ be
\begin{equation}
    H=U\Sigma V^T.
\end{equation}
The rigid transformation is computed as
\begin{equation}
    R=V\operatorname{diag}(1,1,\det(VU^T))U^T,
\end{equation}
\begin{equation}
    t=\bar{q}-R\bar{p}.
\end{equation}

\subsubsection{\uline{Warping Function}: $\M{warping(\cdot,\cdot,\cdot)}$}
After obtaining $(R,t)$, the rigid warping of source points is given by
\begin{equation}
    W(\hat{p}_i)=R\hat{p}_i+t.
\end{equation}

In this paper, we use rigid warping for both rigid and deformable registration as a simple proof of concept, because the deformation between adjacent frames in deformable scenes is often moderate. Given predicted correspondences and associated local rigid assumptions, we can further interpolate sparse flows for points in $\hat{P}$ using nearest-neighbor interpolation. More sophisticated deformation models, such as embedded deformation graphs\cite{sumner2007embedded,igarashi2005rigid}, can also be integrated\cite{li2022non,qin2023deep}, which we leave for future work.

\subsubsection{\uline{Denoising Transformer}: $f_\theta(\cdot,\cdot,\cdot,\cdot,\cdot,\cdot)$}\label{denoising_transformer}
We empirically find that a simple denoising network is sufficient. Therefore, we employ a lightweight Transformer\cite{vaswani2017attention} as the denoising module. Specifically, we use a 6-layer interleaved attention transformer to update point features during reverse sampling.

\textbf{Attention Layer.}
Following\cite{li2022lepard}, we incorporate rotary positional encoding $\Theta(\cdot)$ into the attention layers. In self-attention, the queries, keys, and values are computed as
\begin{equation}
    q_i=\Theta(p_i)W_q f^{\hat{p}_i},
    \qquad
    k_j=\Theta(p_j)W_k f^{\hat{p}_j},
    \qquad
    v_j=W_v f^{\hat{p}_j},
    \label{att_12}
\end{equation}
and the feature update is
\begin{equation}
    f^{\hat{p}_i}=f^{\hat{p}_i}+MLP\!\left(\operatorname{cat}\left[f^{\hat{p}_i},\sum_j \alpha_{ij}v_j\right]\right),
\end{equation}
where $W_q,W_k,W_v \in \mathbb{R}^{d \times d}$ are learnable matrices, and
\begin{equation}
    \alpha_{ij}=\operatorname{softmax}\left(\frac{q_i k_j^T}{\sqrt{d}}\right).
\end{equation}
Here $MLP(\cdot)$ is a three-layer fully connected network, and $\operatorname{cat}[\cdot,\cdot]$ denotes concatenation. The cross-attention layer has the standard form, where the query is computed from the source and the key-value pair is computed from the target.

\subsubsection{\uline{Matching Function}: $\M{matching\_logits(\cdot,\cdot,\cdot,\cdot)}$}\label{mlogits}
Following\cite{li2022lepard}, we compute position-enhanced matching logits between $\hat{P}$ and $\hat{Q}$ using their features:
\begin{equation}
    \tilde{E}(i,j)=\frac{1}{\sqrt{d}}
    \left\langle
    \Theta(p_i)W_P f^{\hat{p}_i},\,
    \Theta(q_j)W_Q f^{\hat{q}_j}
    \right\rangle.
    \label{init_matching}
\end{equation}
Here $W_P$ and $W_Q$ are learnable matrices. The resulting matching logits are then projected into doubly stochastic matrix space through Sinkhorn normalization\cite{cuturi2013sinkhorn}.

Since KPConv\cite{thomas2019kpconv} is translation-invariant, we employ rotary positional encoding $\Theta(\cdot)$, as defined in\cite{li2022lepard}, to alleviate ambiguities caused by symmetry and repeated structures. This encoding is used in both the denoising transformer and the matching function.

For clarity, the overall denoising procedure is summarized in Algorithm~\ref{denoising_algorithm}.

\begin{algorithm}
\caption{Denoising Module $g_\theta$}
\label{denoising_algorithm}
\begin{algorithmic}[1]
\Require Sampled matching matrix $E^t \in \mathbb{R}^{N\times M}$; point clouds $\hat{P}, \hat{Q}$; associated point features $F^{\hat{P}}, F^{\hat{Q}}$.
\Ensure Target matching matrix $\hat{E}_0$.
\Function{$g_{\theta}$}{$E^t, t, \hat{P}, \hat{Q}, F^{\hat{P}}, F^{\hat{Q}}$}
        \State $\tilde{E}_t \gets \M{f_{sinkhorn}}(E^t)$
        \State $\hat{R}_t, \hat{t}_t \gets \M{soft\_procrustes}(\tilde{E}_t, \hat{P}, \hat{Q})$
        \State $\hat{P}_t \gets \M{warping}(\hat{P}, \hat{R}_t, \hat{t}_t)$
        \State $\tilde{F}^{\hat{P}_t}, \tilde{F}^{\hat{Q}} \gets f_{\theta}(\hat{P}_t, \hat{Q}, F^{\hat{P}}, F^{\hat{Q}}, \Theta(\hat{P}_t), \Theta(\hat{Q}), t)$
        \State $\tilde{E}_0 \gets \M{matching\_logits}(\tilde{F}^{\hat{P}_t}, \tilde{F}^{\hat{Q}}, \Theta(\hat{P}_t), \Theta(\hat{Q}))$
        \State $\hat{E}_0 \gets \M{f_{sinkhorn}}(\tilde{E}_0)$
        \State \Return $\hat{E}_0$
\EndFunction
\end{algorithmic}
\end{algorithm}

\begin{algorithm}
\caption{Training Diff-PCR}
\label{training-diff-pcr}
\begin{algorithmic}[1]
\Require Point clouds $\hat{P}$ and $\hat{Q}$, and associated point features $F^{\hat{P}}, F^{\hat{Q}}$.
\While{not converged}
    \State Sample $E^0$ from ground-truth correspondence supervision
    \State Sample $t \sim \text{Uniform}(1,\ldots,T)$
    \State Sample $\epsilon \sim \mathcal{N}(0,1)^{N \times M}$
    \State $E^t \gets \sqrt{\bar{\alpha}_t}E^0+\sqrt{1-\bar{\alpha}_t}\epsilon$
    \State $\hat{E}_0 \gets g_\theta(E^t,t,\hat{P},\hat{Q},F^{\hat{P}},F^{\hat{Q}})$
    \State Optimize $\mathcal{L}_{simple} = \ell(\hat{E}_0, E^0)$
\EndWhile
\end{algorithmic}
\end{algorithm}

\begin{algorithm}
\caption{Sampling by Diff-PCR}
\label{sampling-diff-pcr}
\begin{algorithmic}[1]
\Require Initial matching matrix $E^T$ from the backbone or white noise; point clouds $\hat{P},\hat{Q}$ and associated point features $F^{\hat{P}},F^{\hat{Q}}$.
\Ensure Target matching matrix $E^0$.
\For{$t=T,\ldots,1$}
    \State $z_t \sim \mathcal{N}(0,1)^{N \times M}$ if $t>1$, otherwise $z_t \gets 0^{N \times M}$
    \State $\hat{E}_0 \gets g_\theta(E^t,t,\hat{P},\hat{Q},F^{\hat{P}},F^{\hat{Q}})$
    \State $\hat{\epsilon}_t \gets \dfrac{E^t-\sqrt{\bar{\alpha}_t}\hat{E}_0}{\sqrt{1-\bar{\alpha}_t}}$
    \State $\sigma_t \gets \eta\sqrt{\dfrac{1-\bar{\alpha}_{t-1}}{1-\bar{\alpha}_t}}\sqrt{1-\dfrac{\bar{\alpha}_t}{\bar{\alpha}_{t-1}}}$
    \State $E^{t-1} \gets \sqrt{\bar{\alpha}_{t-1}}\hat{E}_0 + \sqrt{1-\bar{\alpha}_{t-1}-\sigma_t^2}\hat{\epsilon}_t + \sigma_t z_t$
\EndFor
\State \Return $\hat{E}_0$
\end{algorithmic}
\end{algorithm}

\subsection{Framework Training}
Our framework first uses a KPConv\cite{thomas2019kpconv} branch to generate the downsampled superpoints $\hat{P}$ and $\hat{Q}$ and their associated features $F^{\hat{P}}$ and $F^{\hat{Q}}$. We then use the repositioning transformer in Lepard\cite{li2022lepard} to predict the matching matrix and rigid transformation $(R,t)$. These predictions are supervised by the matching loss $L_M$ and warping loss $L_W$ from\cite{li2022lepard}.

To train the denoising module, we use the Gaussian transition kernel to perturb the ground-truth matching matrix $E^0$ and obtain a noisy matrix $E^t$ at timestep $t$. The denoising module $g_\theta$ is optimized using focal loss, which implements the practical surrogate objective in Eqn.~(\ref{L_simple_new}). The total loss is
\begin{equation}
   L = L_M + L_W + \mathcal{L}_{simple}.
\end{equation}

\begin{figure*}[htbp]
      \centering
      \includegraphics[width=\textwidth]{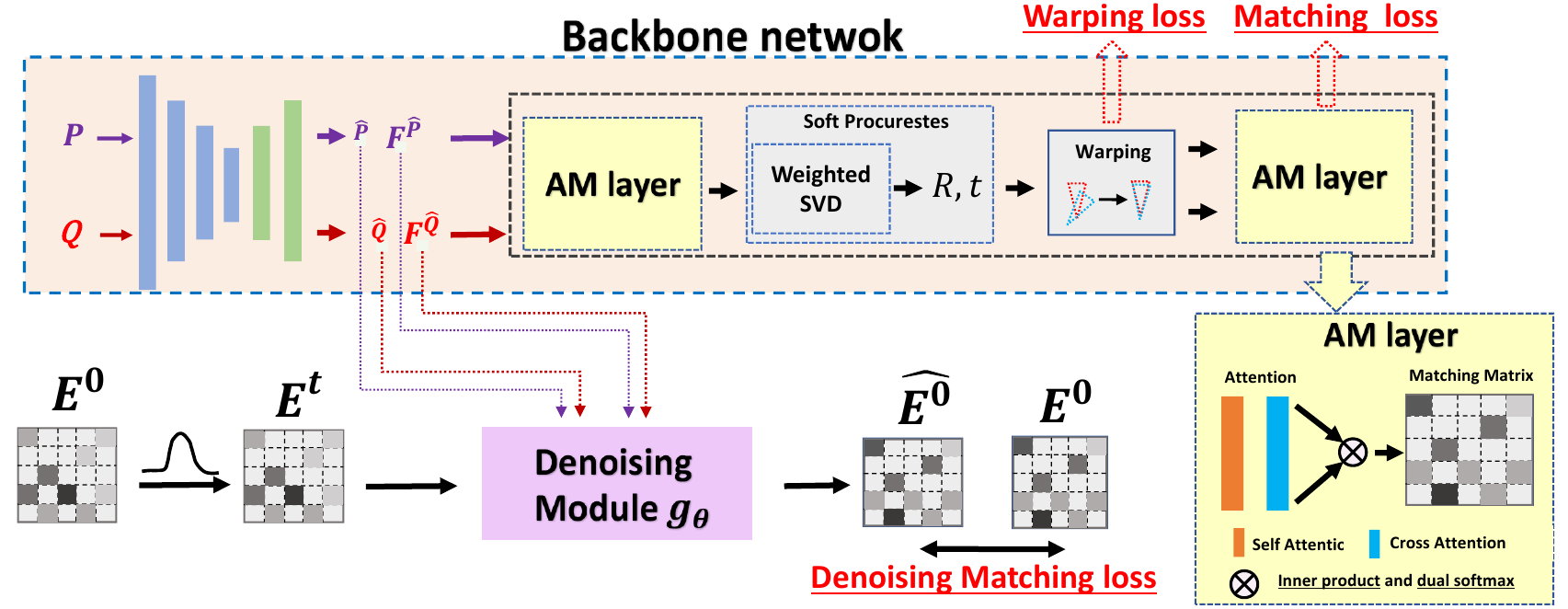}
      \caption{Overview of the training procedure. The framework consists of a KPConv\cite{thomas2019kpconv} backbone and a denoising module. The training procedure of the denoising module is summarized in Algorithm~\ref{training-diff-pcr}. We implement the denoising matching loss by focal loss. After training, only the KPConv backbone and denoising module are retained for reverse sampling, while auxiliary training components are discarded.}
      \label{framework_training}
\end{figure*}

\section{Experiments}

\subsection{Implementation Details}
For the backbone network, we follow Lepard\cite{li2022lepard}. The feature dimension of superpoints $F^{\hat{P}}$ and $F^{\hat{Q}}$ is set to $d=528$. Following\cite{li2022lepard}, we also employ rotary positional encoding in both the denoising transformer and the matching function. Our model is implemented in PyTorch and trained on a single NVIDIA RTX 3090 GPU. We train the model for approximately 30 epochs on both 3DMatch and 4DMatch with a batch size of 2. For 3DMatch and 4DMatch, we follow the training, validation, and testing splits used in Predator\cite{huang2021predator} and Lepard\cite{li2022lepard}, respectively. During inference, we perform 20 reverse denoising steps, while the total number of diffusion steps is set to 1000.

\subsection{Rigid Datasets: 3DMatch and 3DLoMatch}

\subsubsection{Datasets}
3DMatch\cite{zeng20173dmatch} is a widely used indoor benchmark for 3D matching and registration. Following\cite{huang2021predator,qin2022geometric,li2022lepard}, we split the dataset into 46 scenes for training, 8 for validation, and 8 for testing. The overlap ratios in 3DMatch and 3DLoMatch are approximately $>30\%$ and $10\%-30\%$, respectively.

\subsubsection{Metrics}
Following\cite{huang2021predator,qin2022geometric,lee2011hyper}, we use three evaluation metrics:
(1) \textbf{Inlier Ratio (IR)}: the proportion of predicted correspondences whose transformed distance is below a threshold of $0.1\,\mathrm{m}$.
(2) \textbf{Feature Matching Recall (FMR)}: the percentage of scan pairs whose inlier ratio exceeds $5\%$.
(3) \textbf{Registration Recall (RR)}: the percentage of successfully registered pairs whose transformation error satisfies the benchmark criterion, for example RMSE $< 0.2$.

\subsubsection{Results}
\begin{figure}
    \centering
    \captionsetup[subfloat]{labelfont=scriptsize,textfont=scriptsize}
    \subfloat[Lepard~\cite{li2022lepard}]{\includegraphics[width=0.45\linewidth,height=3.0cm]{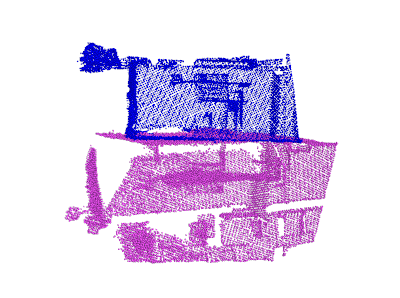}}\quad
    \subfloat[Ours]{\includegraphics[width=0.45\linewidth,height=3.0cm]{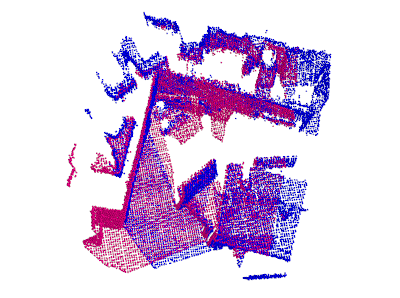}}\quad

    \subfloat[Lepard~\cite{li2022lepard}]{\includegraphics[width=0.45\linewidth,height=3.0cm]{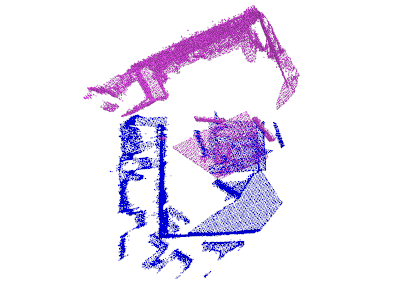}}\quad
    \subfloat[Ours]{\includegraphics[width=0.47\linewidth,height=3.2cm]{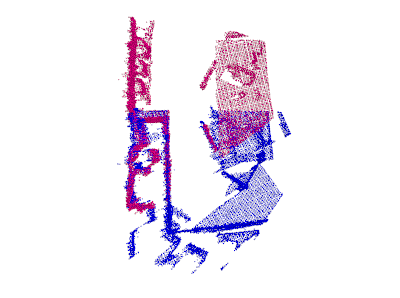}}\quad

    \caption{Qualitative results on the 3DMatch and 3DLoMatch benchmarks. Zoom in for details.}
    \label{ablay_3dmatch}
\end{figure}

We compare our method with several state-of-the-art feature matching methods, including FCGF\cite{choy2019fully}, D3Feat\cite{bai2020d3feat}, Predator\cite{huang2021predator}, and Lepard\cite{li2022lepard}. For fairness, we use RANSAC to estimate the final rigid transformation for all methods. As shown in Table~\ref{tab.3dmatch}, our method achieves the best registration recall. Figure~\ref{ablay_3dmatch} further shows that our method provides more reliable candidate correspondences for the RANSAC process.

\begin{table}
\caption{Quantitative results on the 3DMatch and 3DLoMatch benchmarks. The best results are highlighted in bold, and the second-best results are underlined.}
\centering
\resizebox{\linewidth}{!}{
\begin{tabular}{c|ccc|ccc}
\toprule
\multirow{2}{*}{Method} & \multicolumn{3}{c|}{3DMatch} & \multicolumn{3}{c}{3DLoMatch} \\
 & FMR & IR & RR & FMR & IR & RR \\
\midrule
FCGF\cite{choy2019fully} & 95.20 & 56.90 & 88.20 & 60.90 & 21.40 & 45.80 \\
D3Feat\cite{bai2020d3feat} & 95.80 & 39.00 & 85.80 & 69.30 & 13.20 & 40.20 \\
Predator\cite{huang2021predator} & 96.70 & 58.00 & 91.80 & 78.60 & 26.70 & 62.40 \\
Lepard\cite{li2022lepard} & 97.95 & 57.61 & \uline{93.90} & 84.22 & 27.83 & 70.63 \\
GeoTR\cite{qin2022geometric} & \textbf{98.1} & \uline{72.7} & 92.3 & \uline{88.7} & \uline{44.7} & \textbf{75.4} \\
RoITr\cite{yu2023rotation} & \uline{98.0} & \textbf{82.6} & 91.9 & \textbf{89.6} & \textbf{54.3} & \uline{74.8} \\
Ours (Diff-PCR) & 97.41 & 55.61 & \textbf{94.25} & 80.59 & 22.54 & 73.39 \\
\bottomrule
\end{tabular}}
\label{tab.3dmatch}
\end{table}

\begin{table}
\caption{Quantitative results on the 4DMatch and 4DLoMatch benchmarks. The best results are highlighted in bold, and the second-best results are underlined. S indicates whether the method is supervised.}
\centering
\resizebox{\linewidth}{!}{
\begin{tabular}{c|c|c|cc|cc}
\toprule
\multirow{2}{*}{Category} & \multirow{2}{*}{Method} & \multirow{2}{*}{S} & \multicolumn{2}{c|}{4DMatch} & \multicolumn{2}{c}{4DLoMatch} \\
 & & & NFMR & IR & NFMR & IR \\
\midrule
\multirow{3}{*}{Scene Flow}
& PointPWC\cite{wu2019pointpwc} & \checkmark & 21.6 & 20.0 & 10.0 & 7.2 \\
& FLOT\cite{puy2020flot} & \checkmark & 27.1 & 24.9 & 15.2 & 10.7 \\
& NSFP\cite{li2021neural} &  & 18.5 & 16.3 & 1.2 & 0.5 \\
\midrule
\multirow{5}{*}{Feature Matching}
& D3Feat\cite{bai2020d3feat} & \checkmark & 55.5 & 54.7 & 27.4 & 21.5 \\
& Predator\cite{huang2021predator} & \checkmark & 56.4 & 60.4 & 32.1 & 27.5 \\
& Lepard\cite{li2022lepard} & \checkmark & \uline{83.60} & 82.64 & 66.63 & 55.55 \\
& GeoTR\cite{qin2022geometric} & \checkmark & 83.2 & 82.2 & 65.4 & 63.6 \\
& RoITr\cite{yu2023rotation} & \checkmark & 83.0 & \uline{84.4} & \uline{69.4} & \uline{67.6} \\
& Ours (Diff-PCR) & \checkmark & \textbf{88.39} & \textbf{86.40} & \textbf{76.23} & \textbf{67.80} \\
\bottomrule
\end{tabular}}
\label{tab_4dmatch}
\end{table}

\begin{table}
\caption{Ablation study on initialization for reverse sampling. $E^T_{Backbone}$ denotes initialization from the backbone output. $E^T_{Gaussian}$ denotes initialization from Gaussian white noise $\mathcal{N}(0,1)^{N \times M}$.}
\renewcommand{\arraystretch}{1.2}
\centering
\resizebox{\linewidth}{!}{
\begin{tabular}{c|ccc|ccc}
\toprule
& \multicolumn{3}{c|}{3DMatch} & \multicolumn{3}{c}{3DLoMatch} \\
& FMR & IR & RR & FMR & IR & RR \\
\midrule
$E^T_{Backbone}$ & 97.23 & 55.61 & 94.23 & 83.01 & 23.69 & 72.55 \\
$E^T_{Gaussian}$ & 97.41 & 55.61 & 94.25 & 80.59 & 22.54 & 73.39 \\
\midrule
& \multicolumn{3}{c|}{4DMatch} & \multicolumn{3}{c}{4DLoMatch} \\
& NFMR & IR &  & NFMR & IR &  \\
\midrule
$E^T_{Backbone}$ & 88.38 & 86.38 &  & 75.94 & 67.64 &  \\
$E^T_{Gaussian}$ & 88.40 & 86.40 &  & 76.09 & 67.73 &  \\
\bottomrule
\end{tabular}}
\label{result_backbone}
\end{table}

\begin{table}
\caption{Results of deterministic DDIM reverse sampling, that is, $z_t=0$ in Eqn.~(\ref{ddim_sampling_new}). Backbone denotes initialization from the backbone output. Gaussian denotes initialization from white noise $\mathcal{N}(0,1)^{N \times M}$.}
\renewcommand{\arraystretch}{1.2}
\centering
\resizebox{\linewidth}{!}{
\begin{tabular}{c|ccc|ccc}
\toprule
\multirow{2}{*}{} & \multicolumn{3}{c|}{3DMatch} & \multicolumn{3}{c}{3DLoMatch} \\
 & FMR & IR & RR & FMR & IR & RR \\
\midrule
$E^T_{Gaussian}$ & 97.66 & 58.86 & 94.27 & 83.79 & 27.26 & 73.80 \\
$E^T_{Backbone}$ & 97.66 & 58.82 & 94.13 & 83.79 & 27.26 & 73.89 \\
\midrule
\multirow{2}{*}{} & \multicolumn{3}{c|}{4DMatch} & \multicolumn{3}{c}{4DLoMatch} \\
 & NFMR & IR &  & NFMR & IR &  \\
\midrule
$E^T_{Gaussian}$ & 88.34 & 86.36 &  & 76.22 & 67.82 &  \\
$E^T_{Backbone}$ & 88.72 & 86.72 &  & 76.48 & 68.16 &  \\
\bottomrule
\end{tabular}}
\label{deter_sampling}
\end{table}

\begin{table}
\caption{Non-rigid registration results on 4DMatch-F and 4DLoMatch-F. Using correspondences generated by our method, we perform non-rigid registration with NDP\cite{li2022non}.}
\centering
\resizebox{\linewidth}{!}{
\begin{tabular}{c|cccc|cccc}
\toprule
\multirow{2}{*}{Method} & \multicolumn{4}{c|}{4DMatch-F} & \multicolumn{4}{c}{4DLoMatch-F} \\
 & EPE$\downarrow$ & AccS$\uparrow$ & AccR$\uparrow$ & Outlier$\downarrow$ & EPE$\downarrow$ & AccS$\uparrow$ & AccR$\uparrow$ & Outlier$\downarrow$ \\
\midrule
PointPWC\cite{wu2019pointpwc} & 0.182 & 6.25 & 21.49 & 52.07 & 0.279 & 1.69 & 8.15 & 55.70 \\
FLOT\cite{puy2020flot} & 0.133 & 7.66 & 27.15 & 40.49 & 0.210 & 2.73 & 13.08 & 42.51 \\
GeomFmaps [9] & 0.152 & 12.34 & 32.56 & 37.90 & 0.148 & 1.85 & 6.51 & 64.63 \\
Synorim-pw [19] & 0.099 & 22.91 & 49.86 & 26.01 & 0.170 & 10.55 & 30.17 & \uline{31.12} \\
Lepard+NICP\cite{li2022lepard} & 0.097 & 51.93 & 65.32 & 23.02 & 0.283 & 16.80 & 26.39 & 52.99 \\
Lepard+NDP\cite{li2022non} & \uline{0.075} & \uline{62.85} & \uline{75.26} & \uline{16.78} & \uline{0.169} & \uline{28.65} & \uline{43.37} & 32.14 \\
Ours (Diff-PCR)+NDP & \textbf{0.062} & \textbf{65.52} & \textbf{78.75} & \textbf{13.84} & \textbf{0.141} & \textbf{32.29} & \textbf{48.96} & \textbf{25.75} \\
\bottomrule
\end{tabular}}
\label{non_rigid_registration}
\end{table}

\subsection{Non-Rigid Datasets: 4DMatch and 4DLoMatch}

\subsubsection{Datasets}
4DMatch and 4DLoMatch\cite{li2022lepard} are benchmarks generated from animation sequences in DeformingThings4D\cite{li20214dcomplete}. We follow the dataset split provided in\cite{li2022lepard}. The overlap ratio ranges from $45\%$ to $92\%$ in 4DMatch and from $15\%$ to $45\%$ in 4DLoMatch.

\subsubsection{Metrics}
Following Lepard\cite{li2022lepard}, we use two evaluation metrics to assess the quality of predicted matches.

\textbf{Inlier Ratio (IR).}
This metric measures the proportion of correct correspondences in the predicted correspondence set $\mathcal{K}_{pred}$:
\begin{equation}
    IR = \frac{1}{|\mathcal{K}_{pred}|}\sum_{(p,q)\in \mathcal{K}_{pred}}[\|W_{gt}(p)-q\|_2<\sigma],
\end{equation}
where $W_{gt}(\cdot)$ is the ground-truth warping function, $[\cdot]$ denotes the Iverson bracket, and $\sigma=0.04\,\mathrm{m}$.

\textbf{Non-rigid Feature Matching Recall (NFMR).}
This metric measures the fraction of ground-truth correspondences $(u,v)\in \mathcal{K}_{gt}$ that can be recovered from $\mathcal{K}_{pred}$. We first construct the predicted source set $\mathcal{A}=\{p\mid (p,q)\in \mathcal{K}_{pred}\}$ and the associated sparse flow field $\mathcal{F}=\{q-p\mid (p,q)\in \mathcal{K}_{pred}\}$. For any source point $u \in \mathcal{K}_{gt}$, the flow at $u$ is recovered by inverse-distance interpolation:
\begin{equation}
     \Gamma(u,\mathcal{A},\mathcal{F}) =
     \sum_{\mathcal{A}_i \in knn(u,\mathcal{A})}
     \frac{\mathcal{F}_i\|u-\mathcal{A}_i\|_2^{-1}}
     {\sum_{\mathcal{A}_i \in knn(u,\mathcal{A})}\|u-\mathcal{A}_i\|_2^{-1}},
\end{equation}
where $knn(\cdot,\cdot)$ denotes $k$-nearest neighbors search with $k=3$. NFMR is then defined as
\begin{equation}
    NFMR=\frac{1}{|\mathcal{K}_{gt}|}\sum_{(u,v)\in\mathcal{K}_{gt}}[\|\Gamma(u,\mathcal{A},\mathcal{F})-v\|_2<\sigma].
\end{equation}

\begin{figure*}
    \centering
    \subfloat{\includegraphics[width=0.23\linewidth,height=3cm]{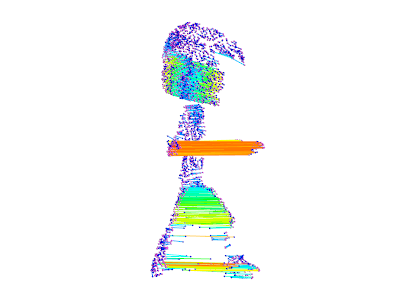}}
    \hfill
    \subfloat{\includegraphics[width=0.23\linewidth,height=3cm]{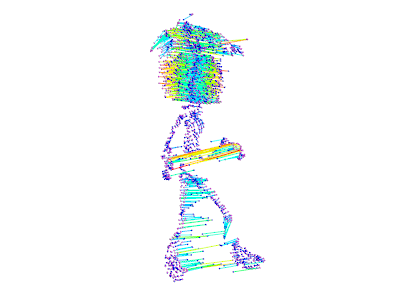}}
    \hfill
    \subfloat{\includegraphics[width=0.23\linewidth,height=3cm]{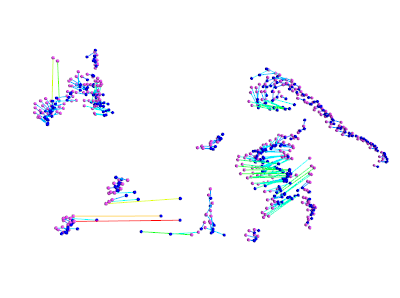}}
    \hfill
    \subfloat{\includegraphics[width=0.23\linewidth,height=3cm]{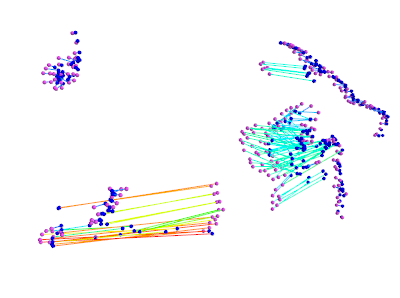}}
    
    \subfloat{\includegraphics[width=0.23\linewidth,height=3cm]{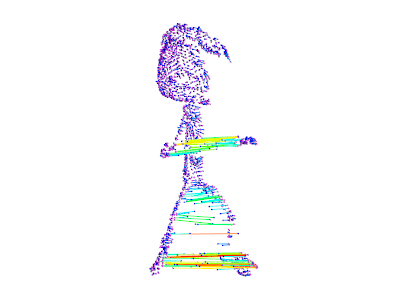}}
    \hfill
    \subfloat{\includegraphics[width=0.23\linewidth,height=3cm]{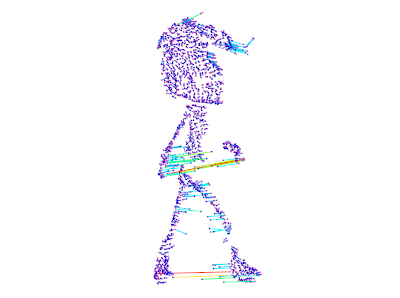}}
    \hfill
    \subfloat{\includegraphics[width=0.23\linewidth,height=3cm]{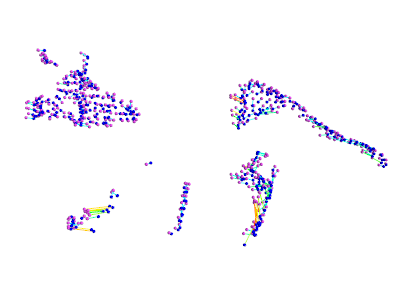}}
    \hfill
    \subfloat{\includegraphics[width=0.23\linewidth,height=3cm]{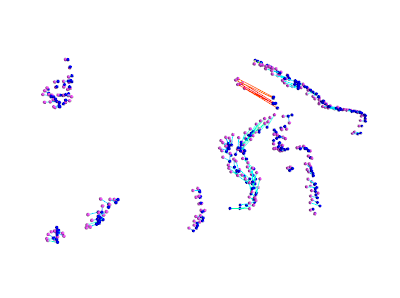}}
    
    \caption{Qualitative results on the 4DMatch and 4DLoMatch benchmarks. The first row shows Lepard\cite{li2022lepard}, and the second row shows our method. Red and green indicate matching errors in two directions. Zoom in for details.}
    \label{visual_4dmatch}
    \vspace{-0.3cm}
\end{figure*}

\begin{figure*}
    \centering
    \subfloat[Step = 0]{\includegraphics[width=0.20\linewidth,height=3cm]{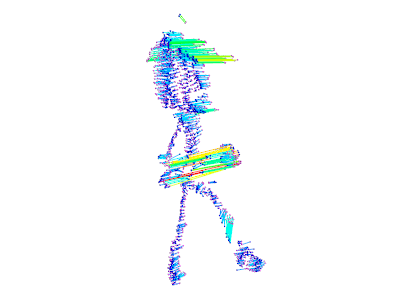}}
    \hfill
    \subfloat[Step = 5]{\includegraphics[width=0.21\linewidth,height=3.1cm]{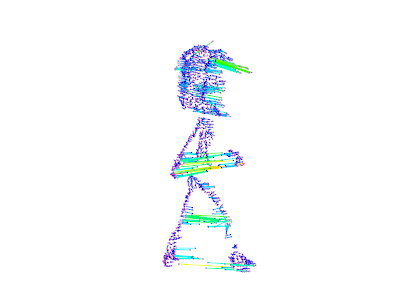}}
    \hfill
    \subfloat[Step = 10]{\includegraphics[width=0.17\linewidth,height=2.7cm]{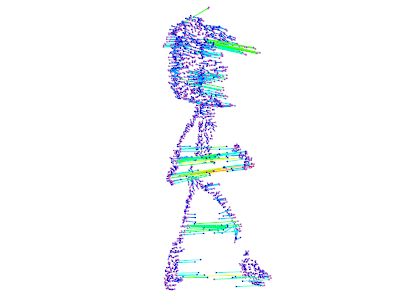}}
    \hfill
    \subfloat[Step = 15]{\includegraphics[width=0.20\linewidth,height=3cm]{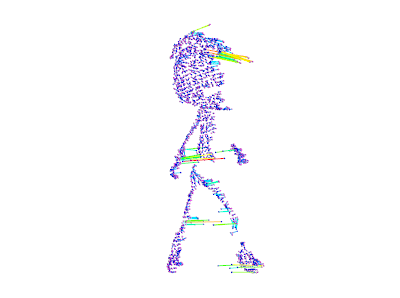}}
    \hfill
    \subfloat[Step = 20]{\includegraphics[width=0.20\linewidth,height=3cm]{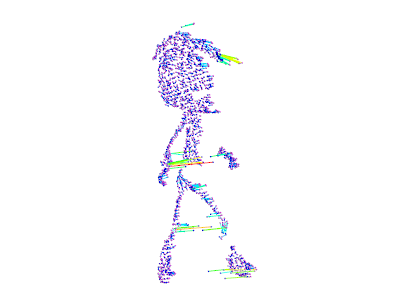}}
    \caption{An example of the reverse sampling process. Red and green indicate matching errors in two directions. Zoom in for details.}
    \label{sampling_steps}
    \vspace{-0.3cm}
\end{figure*}

\subsubsection{Results}
We compare our method with two categories of state-of-the-art methods. The first category includes scene flow methods such as PointPWC\cite{wu2019pointpwc}, FLOT\cite{puy2020flot}, and NSFP\cite{li2021neural}. The second category includes feature matching methods such as D3Feat\cite{bai2020d3feat}, Predator\cite{huang2021predator}, and Lepard\cite{li2022lepard}. As shown in Table~\ref{tab_4dmatch}, our method achieves substantial improvements over the baseline Lepard\cite{li2022lepard}. We also visualize representative examples in Fig.~\ref{visual_4dmatch}. The results show that the proposed denoising optimizer effectively reduces matching errors under different overlap ratios.

\begin{figure*}
    \centering
    \begin{minipage}{\textwidth}
        \centering
        \includegraphics[width=0.23\textwidth,height=3cm]{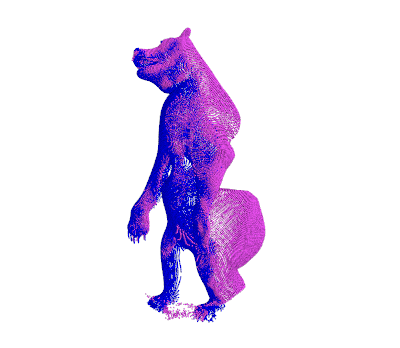}
        \hfill
        \includegraphics[width=0.23\textwidth,height=3cm]{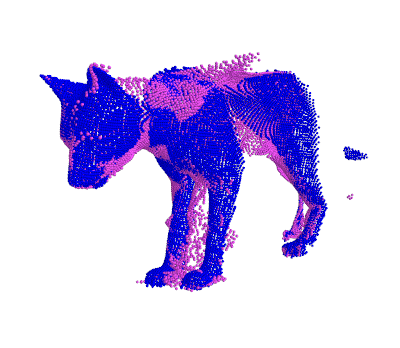}
        \hfill
        \includegraphics[width=0.23\textwidth,height=3cm]{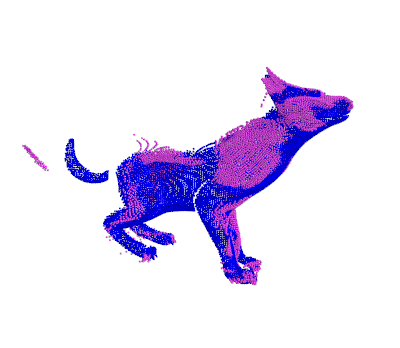}
        \hfill
        \includegraphics[width=0.23\textwidth,height=3cm]{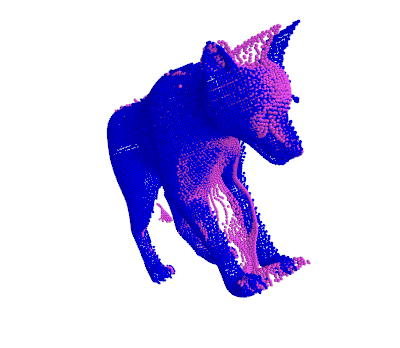}
    \end{minipage}
    
    \begin{minipage}{\textwidth}
        \centering
        \includegraphics[width=0.23\textwidth,height=3cm]{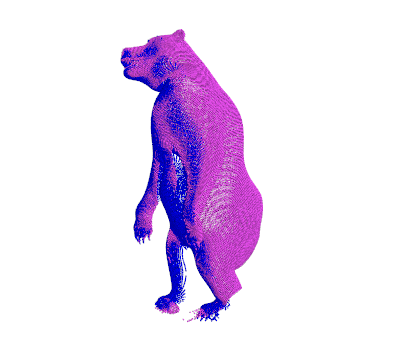}
        \hfill
        \includegraphics[width=0.23\textwidth,height=3cm]{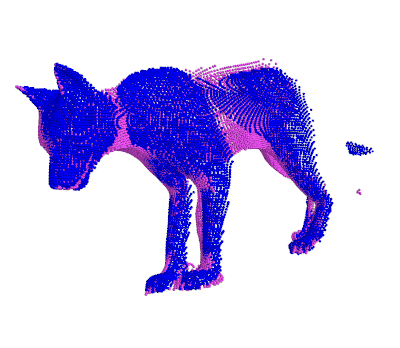}
        \hfill
        \includegraphics[width=0.23\textwidth,height=3cm]{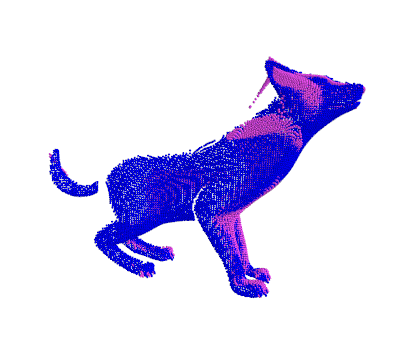}
        \hfill
        \includegraphics[width=0.23\textwidth,height=3cm]{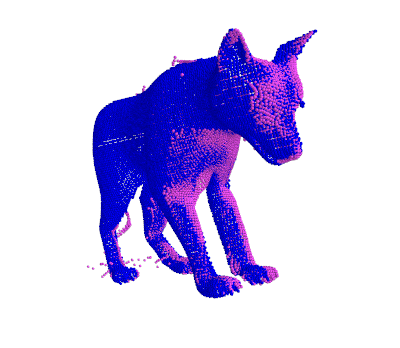}
    \end{minipage}

    \caption{Qualitative results of non-rigid registration on 4DMatch and 4DLoMatch. The first row is generated by Lepard+NDP\cite{li2022non}, and the second row is generated by Ours+NDP\cite{li2022non}. Zoom in for details.}
    \label{non_rigid_registration_vis}
\end{figure*}

\subsection{Ablation Studies and Discussion}

\subsubsection{Denoising Transformer Design}
We empirically observe that combining KPConv\cite{thomas2019kpconv} with a simple transformer already yields strong improvements over several state-of-the-art registration methods. Recent studies have shown that more powerful feature embedding networks can further improve denoising performance. In principle, our framework can benefit from stronger transformers or from additional semantic and geometric priors\cite{qin2022geometric,yu2023peal}. For example, for rigid registration, GeoTR\cite{qin2022geometric} or the semantic-enhanced geometric transformer\cite{wu2023sgfeat} could be incorporated into $f_\theta$. We leave this extension to future work. In the current version, we use a standard attention layer with rotary positional encoding\cite{li2022lepard}, which works for both rigid and non-rigid scenarios.

\subsubsection{Reverse Sampling Initialization}
The DDPM framework is designed to remove noise from perturbed samples. In practice, a registration backbone often provides a reasonably good initialization, which can then be refined. To verify that our denoising network learns a meaningful search gradient rather than merely refining one specific initialization, we perform an ablation study with two starting points: a pre-trained backbone prediction and Gaussian white noise. The results in Table~\ref{result_backbone} show that our denoising network performs well under both settings. This indicates that the learned reverse process is robust to initialization and can converge toward a high-quality solution even when starting from white noise.

\begin{table}
\caption{Ablation study on the number of reverse sampling steps. $0^*$ denotes the result of the pretrained Lepard\cite{li2022lepard} model without denoising refinement.}
\tiny
\centering
\resizebox{\linewidth}{!}{
\begin{tabular}{c|cc|cc}
\toprule
\multirow{2}{*}{Iters} & \multicolumn{2}{c|}{RR} & \multicolumn{2}{c}{NFMR/IR} \\
 & 3DMatch & 3DLoMatch & 4DMatch & 4DLoMatch \\
\midrule
0$^*$ & 93.50 & 69.00 & 83.74/82.74 & 66.94/55.74 \\
\midrule
1 & 93.96 & 73.39 & 85.34/83.93 & 73.11/65.26 \\
2 & 93.96 & 73.12 & 85.24/83.84 & 73.27/65.19 \\
3 & 94.04 & 73.52 & 85.52/84.06 & 73.19/65.22 \\
10 & 94.27 & 73.59 & 87.99/86.07 & 75.46/67.15 \\
20 & 94.23 & 73.54 & 88.39/86.64 & 76.16/67.82 \\
50 & 93.98 & 73.35 & 88.61/86.62 & 76.34/68.03 \\
100 & 94.31 & 73.37 & 88.55/86.58 & 76.46/68.15 \\
200 & 94.36 & 73.36 & 88.59/86.61 & 76.42/68.10 \\
\bottomrule
\end{tabular}}
\label{result_iter_num}
\end{table}

\subsubsection{Reverse Sampling Steps}
In our method, the denoising module acts as an optimizer that searches for a better matching matrix. We hypothesize that increasing the number of search steps can improve the final solution. To verify this, we evaluate the method with 1 to 200 reverse sampling steps. As shown in Table~\ref{result_iter_num}, the performance of our method generally improves as the number of steps increases. This behavior differs from that of several iterative refinement methods\cite{li2022lepard}, whose performance often degrades after too many iterations. Even with only one denoising step, our method already improves over the baseline, which demonstrates the effectiveness of the learned optimizer. Figure~\ref{sampling_steps} further illustrates that the matching error gradually decreases as reverse sampling proceeds.

\subsubsection{Deterministic Sampling vs. Stochastic Sampling}
Several empirical studies\cite{chen2023diffusionpcr,jiang2023se} have reported that deterministic sampling can outperform the original stochastic DDPM reverse process. We therefore compare the two strategies. The deterministic results are shown in Table~\ref{deter_sampling}, while the stochastic results are shown in Table~\ref{result_backbone}. We observe that deterministic sampling yields slightly better performance. One possible reason is that the injected Gaussian noise in stochastic sampling does not always lead to the best search trajectory for this task.

\subsection{Integrating Correspondences into Non-rigid Registration}
Given the correspondences predicted by our method, we further perform non-rigid registration to validate their usefulness. Since the 4DMatch benchmark contains many examples dominated by rigid motion, we follow NDP\cite{li2022non} and remove near-rigid cases, resulting in the filtered subsets 4DMatch-F and 4DLoMatch-F. Because our current denoising module uses rigid warping, we evaluate non-rigid registration on these filtered datasets to demonstrate the effectiveness of the present design. We use NDP\cite{li2022non} as the downstream non-rigid registration framework. As shown in Table~\ref{non_rigid_registration}, the correspondences generated by our method consistently improve the performance of deformable registration. Figure~\ref{non_rigid_registration_vis} also shows that our method better handles symmetric structures such as front legs and reduces adhesion artifacts.

\begin{table}
\caption{Comparison of runtime per iterative step, or reverse sampling step.}
\tiny
\centering
\resizebox{\linewidth}{!}{
\begin{tabular}{c|cc|cc}
\toprule
\multirow{2}{*}{Method} & \multicolumn{2}{c|}{3DMatch} & \multicolumn{2}{c}{3DLoMatch} \\
 & RR & Time (sec.) & RR & Time (sec.) \\
\midrule
GeoTR\cite{qin2022geometric} & 92.0 & 0.296 & 74.0 & 0.284 \\
\midrule
GeoTR + PEAL\cite{yu2023peal}, 1 step & 93.7 & 0.663 & \uline{77.8} & 0.642 \\
GeoTR + DiffusionPCR\cite{chen2023diffusionpcr}, 1 step & \uline{93.9} & 0.625 & \textbf{78.2} & \uline{0.620} \\
KPConv + Diff-PCR, 1 step & \textbf{93.96} & \textbf{0.032} & 73.39 & \textbf{0.036} \\
\midrule
GeoTR + PEAL\cite{yu2023peal}, 5 steps & 94.0 & 2.131 & \uline{78.5} & 2.074 \\
GeoTR + DiffusionPCR\cite{chen2023diffusionpcr}, 5 steps & \textbf{94.4} & \uline{1.939} & \textbf{80.0} & \uline{1.964} \\
KPConv + Diff-PCR, 5 steps & \uline{94.13} & \textbf{0.095} & 73.78 & \textbf{0.098} \\
\bottomrule
\end{tabular}}
\label{result_time}
\end{table}

\subsection{Lightweight Design of the Denoising Module}
Because of its lightweight design, our denoising module is substantially faster than existing diffusion-based registration methods. This efficiency enables more denoising iterations within a practical runtime budget. Table~\ref{result_time} reports the runtime per reverse denoising step and compares it with recent methods. Under this lightweight design, our method achieves competitive performance on 3DMatch with significantly lower cost. If GeoTR were used as the denoising transformer, the performance on 3DLoMatch could likely be further improved, although this would also increase the per-step cost. We leave this extension to future work.

\section{Conclusion}
This paper presents a novel framework that applies a diffusion model in doubly stochastic matrix space to learn a search direction for the optimal matching matrix. We further design a lightweight denoising module that greatly reduces the computational cost of iterative reverse sampling. Experimental results on both rigid and non-rigid benchmarks demonstrate the effectiveness and efficiency of the proposed framework.

{\normalem
\bibliographystyle{IEEEtran}
\bibliography{root}
}

\newpage
\vfill

\end{document}